\documentclass[a4paper,conference,10pt]{IEEEtran}
\usepackage{times}
\usepackage{epsfig}
\usepackage{graphicx}
\usepackage{amsmath}
\usepackage{amssymb}
\usepackage{multirow}
\usepackage{balance}
\usepackage{romannum}

\usepackage{ctable}

\ifCLASSINFOpdf
\else
\fi
\begin{document}
%
\title{MDCN: Multi-Scale, Deep Inception Convolutional Neural Networks for Efficient Object Detection}

\author{\IEEEauthorblockN{Wenchi Ma$^{\dag}$, Yuanwei Wu$^{\dag}$, Zongbo Wang$^{\ddag}$, Guanghui Wang$^{\dag}$}
\IEEEauthorblockA{$^{\dag}$Department of Electrical Engineering and Computer Science\\
University of Kansas, Lawrence, Kansas 66045 USA\\
Email: \{wenchima, y262w558, ghwang\}@ku.edu}
\IEEEauthorblockA{$^{\ddag}$Ainstein Inc., 2029 Becker Drive Lawrence, Kansas 66047 USA}}


%


\maketitle

\begin{abstract}
Object detection in challenging situations such as scale variation, occlusion, and truncation depends not only on feature details but also on contextual information. Most previous networks emphasize too much on detailed feature extraction through deeper and wider networks, which may enhance the accuracy of object detection to certain extent. However, the feature details are easily being changed or washed out after passing through complicated filtering structures. To better handle these challenges, the paper proposes a novel framework, multi-scale, deep inception convolutional neural network (MDCN), which focuses on wider and broader object regions by activating feature maps produced in the deep part of the network. Instead of incepting inner layers in the shallow part of the network, multi-scale inceptions are introduced in the deep layers. The proposed framework integrates the contextual information into the learning process through a single-shot network structure. It is computational efficient and avoids the hard training problem of previous macro feature extraction network designed for shallow layers. Extensive experiments demonstrate the effectiveness and superior performance of MDCN over the state-of-the-art models. 
\end{abstract}


%
\IEEEpeerreviewmaketitle

\section{Introduction}
The ability of detailed feature extraction has become one of a common standard for convolutional neural networks (CNNs). Many researchers have tried to improve their feature extraction networks by making them deeper or wider, which is also the most common strategy for difficult object detection tasks such as small objects and occluded ones. Recently, more and more object detection models choose to use the original size of image data with the purpose of obtaining more detail information. However, these methods bring in huge computation burden and it is not the best choice in real-world applications processing large-size image data. 

This paper proposes a novel framework, called MDCN, which covers wide-context receptive fields and extracts multi-scale features by introducing inception modules to deep part of the network. These inception modules consist of multi-scale filtering units whose responsive regions account for a larger proportion of the feature maps produced in the deep part of the network, and they will activate objects with different sizes of background. MDCN maintains a relatively small feature extraction structure and the proposed inception modules only process feature maps with smaller sizes as they are produced later through forward propagation, which enables computational efficiency and better portability over other models.  

\begin{figure*}[h]
	\centering
	\includegraphics[width=1.0\linewidth]{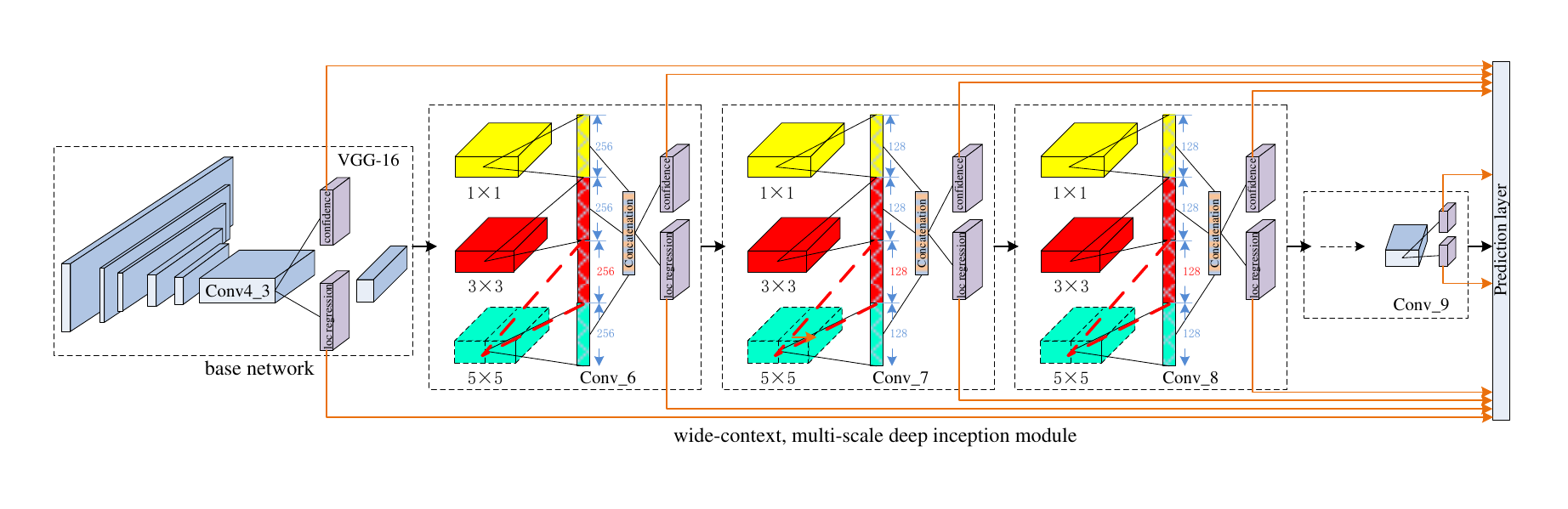}\\%
	\vspace{-8mm}
	\caption{\textbf{The architecture of MDCN.} The red, yellow and green boxes consist of wide-context, multi-scale deep inception structure. Each color denotes one kind of filter size. Purple boxes show classification and localization regression layers according to SSD\cite{liu2016ssd}.}
	\label{fig:detectionrecog1}
	
\end{figure*} 

In principle, MDCN is an intuitive extension of feature extraction, while constructing deep inception modules properly is critical to yield better results. Development in computer hardware enables the training of macro CNNs, which stimulates the research in CNN structures in the direction of deeper and wider for better detailed feature extraction~\cite{erhan2014scalable,he2016deep,zagoruyko2016wide}. The feature extraction ability of networks has been enhanced dramatically from the original LeNet~\cite{lecun1998gradient} with only 5 layers, VGG-16~\cite{russakovsky2015imagenet} to GoogleNet~\cite{szegedy2015going}, residual networks (ResNets) which have surpassed 100 layers~\cite{he2016deep}, wide-residual networks~\cite{zagoruyko2016wide}. For a very deep CNN network, problems of gradient vanishing and feature propagation emerge. The introduction of skip-connection, the propose of Highway networks~\cite{srivastava2015training}, stochastic depth technology~\cite{huang2016deep} and FractalNet~\cite{larsson2016fractalnet} all tried to create shorter paths between earlier and later layers in order to avoid the two problems to certain degree. ResNet-101~\cite{he2016deep}, with skip-connection has become the most popular structure when being used as the base network which has shown its advantage of feature extraction and representation in object detection and segmentation tasks~\cite{fu2017dssd, he2017mask} over other methods. Many researchers tried to adopt ResNet-101 as their main feature extraction network. Mask R-CNN utilizes ResNet-101 as the main body of its network~\cite{he2017mask}. DSSD~\cite{fu2017dssd} achieves its best performance with Residual-101 on PASCAL VOC2007 and PASCAL VOC2012 datasets. However, DSSD only yields 2 percent of increase of mAP (mean average precision) by replacing the original VGG-16~\cite{russakovsky2015imagenet} base network with ResNet-101 while its detection speed decreases from 19 FPS to 6.6 FPS~\cite{fu2017dssd}. Mask R-CNN also suffers from the unsatisfactory detection speed brought by ResNet-101~\cite{he2017mask}. 

The corresponding approach towards being deeper is to increase the width of the networks. The inception module proposed by GoogleNet~\cite{szegedy2015going} enhances the ability of feature extraction of CNNs by concatenating feature-maps produced by filters of different sizes at the same layer level, which reduces the request for the depth of networks. Inception modules are very flexible and portable which can be used in any kind of layer units. Various inception modules have always been the hot research topic like the Inception-v4, inception-ResNet and the combination of residual networks~\cite{szegedy2017inception, szegedy2016rethinking, zagoruyko2016wide}. Residual-inception and its variances~\cite{zagoruyko2016wide} showed their advantage over achievements from individual techniques. While, simply increasing the number of filters in each layer of ResNet is able to improve performances provided the depth is sufficient~\cite{huang2016densely} and the computational load is not the biggest concern. 

The trend of constructing macro networks to enhance the power of feature extraction has been continually challenging the computational power of hardware. Most advanced CNN networks pay high computation cost and consume quite a lot of memory. They have limitations for applications in need of real-time performance and better portability. Most of them share the same key characteristic: tending to make full use of feature maps produced by earlier layers, which have relatively larger sizes with more details. While these approaches do not pay enough attention to contextual information which object detection seriously depends on, especially for those objects with small sizes or being occluded. Furthermore, deepening and widening is also the essential reason that leads to worse efficiency. In this paper, we focus more on contextual information by introducing the multi-scale, deep inception structure. The main contributions of this study include: 

(1) We propose MDCN that puts the extraction of contextual features into a single-shot learning process by incepting multi-scale filtering units in the deep part of the network. 

(2) Information square inception modules are proposed to detect objects with multi-size context expression while maintaining a high computation efficiency by parameter sharing. 

(3) The proposed framework achieves better performance with a relatively shallow network at a real-time speed. The proposed model and trained parameters will be available on the author's website. 

We evaluate MDCN on the prevailing on-road dataset (KITTI)~\cite{geiger2012we} and utilize input images that are scaled to a small size considering the condition of most real-world hardware equipment. We compare the results of our models with other state-of-the-art methods using the same data size. 

\section{MDCN}

MDCN activates wider and broader object regions by wide-context receptive fields and considers multi-resolution features by multi-scale filtering. It integrates the extraction towards various objective contexts into a single-shot learning process without inserting man-made proposals. Moreover, wide-context, multi-scale filtering structures are incepted in deep layers of the network where their output features reflect the most important features of objects and considers the relationships between objects and objects, objects and context. The overall architecture of MDCN is shown in Fig~\ref{fig:detectionrecog1}. It is constituted of the base network (VGG-16), deep detection network provided by the proposed wide-context, multi-scale deep inception modules and the final prediction layer. In each module, we adopt three kinds of filters with different scales to activate features produced by deep layers as labeled by the three colors. The results of the four deep modules are put directly to the final prediction layer as orange arrows indicates. 

\subsection{Detection Pipeline}

MDCN is a single-shot and single-stage detection pipeline in which region proposal, wide-angle contextual information learning and object classification are performed by a single network simultaneously. As shown in Fig.~\ref{fig:detectionrecog1}, the base network first extracts the high-resolution, low-dimensional feature map from the original input image. Then, the feature map is fed into our inception filtering units for the extraction of object main-body features and multi-size contextual information. The proposed wide-context, multi-scale inception structures are designed to the first three levels of the top layers given that the other way for network to learn contextual information that should not be ignored is from previous and later layers~\cite{ren2017accurate}. In order to shorten the path of feature transmission and minimize the probability of features being changed or washed out, the output feature information from the inception units is fed into the final prediction layer directly. MDCN tries to make full use of the impacts of multi-scale features so as to draw all output feature maps from top layers and the feature information with higher resolution from layer conv4\_3. 

The base network served by VGG-16 is pre-trained on ImageNet. MDCN then would be transferred to learn on target dataset. The algorithm of object detection MDCN adopts is multi-box technique, which was proposed in SSD~\cite{liu2016ssd}, one of the state-of-the-art object detectors that realizes high-precision detection and maintains a real-time speed. Multi-box technique discretizes its output space of bounding boxes into a set of default boxes over various aspects ratios and scales for every feature map location and it realizes classification and localization by bounding box regression with multi-scale feature information from continuous extraction units. For MDCN, it assigns a set of default bounding boxes for each feature map cell where the position of each box instance relative to its corresponding cell is fixed. Specifically, in each feature map cell, the offsets relative to the default boxes and the scores of every class indicating the existence of class instance in each box are predicted. Given $k$ boxes to each given location, we calculate $c$ class scores and 4 offsets relative to the default box. This leads to a total of $k(c+4)$ filters working for each location inside a feature maps. Thus, for a feature map with the dimension of $x\times y$, its number of outputs should be $k(c+4)xy$. This kind of multi-box technique efficiently discretizes the space of possible output box shapes, which in turn enhances the accuracy of object localization and classification. 

\subsection{Wide-Context Receptive Field}
\subsubsection{Contextual Information}
Deep learning based object detection tends to realize more practical tasks. Some specific datasets, made from certain scenes, have become popular benchmarks, like KITTI~\cite{geiger2012we}. They provide more challenging labeled objects that are smaller and severely occluded. Many models, in this situation, are hard to achieve effective detection. For these tasks, researchers have found solutions from objects themselves. It is found the detection towards difficult objects is not merely hinged on detailed features but also contextual information as details are not sufficient for objects with small sizes or being occluded~\cite{torralba2003contextual,sui2018correlation}. 

Many methods exploit certain number of contextual regions centered on each object. Sermanet $et\ al.$~\cite{sermanet2013pedestrian} used two contextual regions centered on each object for pedestrian detection. Later, some researchers proposed to
use 10 contextual regions around each object with different crops~\cite{gidaris2015object}. The multipath network is proposed with the purpose of better detecting objects with various sizes by filtering ROI data with different scene sizes in several network paths~\cite{zagoruyko2016multipath} in order to improve localization and classification accuracy. He $et\ al.$~\cite{he2014spatial} make full of context by aggregating CNN features prior to classification using different sized pooling regions. Certain sizes of contextual regions around objects are limited by its number and multi-stage or multi-path structures would introduce a great number of parameters and much more computational load. Moreover, these methods, including the one with different sized pooling regions, merely provide more contextual information for detector while they rarely realize the synchronous learning of contextual features. 

\subsubsection{Wide-Context Receptive Field}
MDCN integrates the learning of contextual information in a single-shot network structure in a implicit way and it does not specify certain sizes of contextual regions for objects. Instead, it guides the network to activate various contextual regions by itself during the learning process. This makes feature learning consider contexts in a spontaneous learning process. Wide-context receptive filed covers a larger proportion of a feature map, which would introduce more sensitive activation towards the main-body characteristic of the objects and their relationships with context. In MDCN, wide-context receptive field is hinged on both the feature maps produced in deep layers and the adoption of filters with relatively larger scales in deep layers. Feature maps produced from deep layers are already with relatively small dimensions. Thus, they would cover larger proportion of the original scene, then more contextual information can be involved into actual learning course. This strengthens the propagation of context information across layers. This can be expressed by the following equations. 
\begin{equation}
\Phi_{n}=f_{n}(\Phi_{n-1})=f_{n}(f_{n-1}(...f_{1}(I))) 
\end{equation}
\begin{equation}
\begin{aligned}
\Phi_{m}&=F_{m}(\Phi_{m-1})\\
&=F_{m}(F_{m-1}(...F_{m-k}(\Phi_{n}))), m-k>n
\end{aligned} 
\end{equation}
\begin{equation}
\begin{aligned}
F_{j}=f_{j}(\Phi_{j-1};W_{j}), m-k \leq j \leq m
\end{aligned} 
\end{equation}
where $m-k$ indicates deep layers, and $n$ refers to earlier layers in base network. $\Phi_{n}$ is the corresponding output feature maps of earlier layer $n$. $\Phi_{m}$ is the corresponding output feature maps of top layer $m$. Function $f_{n}$ maps $\Phi_{n-1}$ to its receptive fields outputs. Function $F_{j}$ maps $\Phi_{j-1}$ to its receptive fields outputs in several channels by the function $f_{j}$ under different weights at layer $j$. $I$ stands for the input image.

From the perspective of computational load, our proposed wide-context receptive field still has its advantage. They are realized by deep inception units, which are defined to process feature maps produced in deep layers. The dimensions of these features have been down-sampled several times throughout earlier filtering process. Thus, although inceptions would increase the computation of a single layer, the smaller-sized feature maps would help weaken this influence. 

\subsection{Information-Square Inception Modules}

The proposed wide-context, multi-scale inception module captures direct output feature maps from base network. In each module, we directly output the input feature information from the previous layer by 1x1 filtering. At the meantime, we conduct 3$\times$3 and 5$\times$5 filtering to offer various wide-context receptive fields activating broader object regions on the input feature maps. In reality, we use two series of 3$\times$3 filters to replace the original 5$\times$5 filter so as to minimize the number of parameters~\cite{szegedy2016rethinking}. The operation of a single inception module can be expressed in equation (4). 

\begin{equation}
\begin{aligned}
F_{j}=f_{j}(f_{j}(\Phi_{j-1}))+2*f_{j}(\Phi_{j-1})+\Phi_{j-1},&\\
m-k \leq j \leq m&
\end{aligned} 
\end{equation}
where all characters have same meanings in equation (2) and (3) and $f_{j}$ operates the 3$\times$3 filtering specifically. By this substitution, the 5$\times$5 filtering unit obtains a 18/25 reduction in the amount of parameters. Furthermore, we use parameter-sharing between the 3$\times$3 and 5$\times$5 tunnel by extracting the output from the first 3$\times$3 filter of the 5$\times$5 filtering unit and concatenate it with other parallel outputs of 3$\times$3 filtering unit as explicitly illustrated by the red dashed lines in Fig.~\ref{fig:detectionrecog1}. Thus, the number of output tunnels of the 3$\times$3 filtering is implicitly doubled (we have the coefficient of 2 for $f_{j}(\Phi_{j-1})$) while we only use one set of parameters in this local part as the ratio for these three kinds of filters are 1:1:1. Thus, the proposed inception unit is in fact a kind of information square as shown in equation (5). This module equals to making square of the sum of identity mapping and 3$\times$3 filtering from the perspective of mathematics, while its amount of parameters is reduced. 
\begin{equation}
\begin{aligned}
F_{j}^{2}(\Phi_{j-1}) &= (f_{j}^{2}+2 \times f_{j} +1)(\Phi_{j-1}) \\
&= ((f_{j}+1)^{2})(\Phi_{j-1}), m-k \leq j \leq m
\end{aligned} 
\end{equation}

\begin{table*}[ht]
	\caption{The comparison results of different models in terms of average precision(\%) on KITTI validation set.} 
	\begin{center}
		\def\arraystretch{1.2}
		\scalebox{0.96}[0.96]{			
			\begin{tabular} {|c|c|c|c|c|c|c|c|c|c|c|}
				\hline
				\multirow{2}{*}{Model} 
				& \multicolumn{3}{c|}{Car} & \multicolumn{3}{c|}{Pedestrian} & \multicolumn{3}{c|}{Cyclist} & \multirow{2}{*}{mAP} \\
				\cline{2-10}
				& Easy & Moderate & Hard & Easy & Moderate & Hard & Easy & Moderate & Hard &\\
				\hline
				SSD & 85.00 & 74.00 & 67.00 & 53.00 & 50.00 & 48.00 & 46.00 & 52.00 & 51.00 & 58\\
				\hline
				ResNet-101 & 87.57 & 76.04 & 68.07 & 50.27 & 47.74 & 45.21 & 49.86 & 53.61 & 51.77 & 58.9\\
				\hline
				WRN-16-4 & 90.08 & 76.8 & 68.5 & 52.29 & 47.88 & 45.3 & 47.71 & 50.36 & 49.38 & 58.7\\
				\hline
				WR-Inception & 87.1 & 77.2 & 68.81 & 55.98 & \textbf{52.51} & 48.61 & 52.9 & 54.63 & 52.87 & 61.18\\
				\hline
				WR-Inception-12 & \textbf{90.36} & 78.24 & 71.11 & 53.26 & 51.08 & \textbf{49.54} & 57.02 & 59.28 & 57.39 & 63.03\\
				\hline
				MDCN-I1 & 88.40 & 87.96 & 87.34 & \textbf{56.39} & 50.37 & 48.86 & 71.58 & 72.21 & \textbf{76.82} & 71.91\\
				\hline
				MDCN-I2 & 88.70 & \textbf{88.19} & \textbf{87.91} & 55.02 & 50.21 & 48.28 & \textbf{73.85} & \textbf{72.66} & 74.95 & \textbf{72.30}\\
				\hline
			\end{tabular} %
		}
	\end{center}
\end{table*}

\subsection{Implementation Details}
\subsubsection{Base Network}
VGG-16 as the base network has only 16 convolutional layers where only the convolution and pooling layers are considered. This VGG network is constituted only by continuous 3$\times$3 filtering, which is regarded as one of the most efficient filter size~\cite{iandola2016squeezenet}. The final prediction layer of MDCN combines low-level features from the layer of conv4\_3 at the depth of 13, of which output feature maps have the resolution of 38$\times$38. 

\subsubsection{Wide-Context Inception Architecture Layout}
There are three wide-context, multi-scale convolution units incepted in the deep part of the MDCN, denoted as Conv\_6, Conv\_7, Conv\_8. Each unit is composed of a 1$\times$1 convolutional unit, followed by an  information square inception module, where the first 1$\times$1 convolution can be introduced as bottleneck layer to reduce the number of input feature maps. The overall layouts of our proposed model is specifically described in Fig.~\ref{fig:detectionrecog1}. The three deep convolutional units (Conv\_6, Conv\_7 and Conv\_8) process the feature maps with sizes 19$\times $19, 10$\times$10 and 5$\times$5, respectively. For comparison, we propose two models, called MDCN-I1 and MDCN-I2, where MDCN-I2 has the layout as the description above with the inception modules in all the three deep convolutional units. But for MDCN-I1, we only design the proposed inception modules in the first two deep units (Conv\_6 and Conv\_7), leaving the unit of Conv\_8 with the same layout as Conv\_9 described in Fig.~\ref{fig:detectionrecog1}.  

\section{Experiments}

We empirically demonstrate the effectiveness of MDCN on the widely used KITTI~\cite{geiger2012we} benchmark. We analyzed the object detection accuracy in terms of average precision (AP), object detection efficiency in terms of speed and model size. To fit the need of real-world applications, we scale all the input images to 300$\times$300 and then perform a thorough comparison with the state-of-the-art methods: SSD~\cite{liu2016ssd}, ResNet-101~\cite{he2016deep, lee2017wide}, WRN-16-4~\cite{zagoruyko2016wide,lee2017wide}, WR-Inception~\cite{lee2017wide}, and WR-Inception-12~\cite{lee2017wide}. These methods use multi-box detector and all of them use the input images that are scaled to 300$\times$300. The proposed framework is implemented using caffe~\cite{jia2014caffe}, compiled with cuDNN computational kernels. All of our experiments are conducted on Tesla K40 GPU.

\subsection{Dataset}
KITTI object detection dataset is designed for autonomous driving, containing many challenging objects like small and occluded cars, pedestrians and cyclists. It is obtained by stereo cameras and lidar scanners in highway, rural and urban driving scenes. KITTI contains 7,481 images for training and validation, and 7,518 images for testing, providing around 40,000 object labels classified as easy, moderate, and hard based on how much objects are occluded and truncated. As the ground truth of test set is not publicly available, we evaluate related models (Table 1) on validation set and report their average precision (AP) on it at the three levels of difficulty following the suggestion in~\cite{geiger2012we,xiang2017subcategory}. Our models are trained to detect 3 categories of objects, including car (merged with motors), pedestrian, and cyclist, as the standard practice provided by KITTI. As we rescale input images into 300$\times$300, compared to the original size of 1242$\times$375, the detection difficulty is greatly increased. The Intersection over Union (IoU) for car, pedestrian, and cyclist are all set to 50\% and all compared methods in Table \Romannum{1} follow this rule. IoU is an evaluation metric used to measure the accuracy in object detection, which is defined as the area of union divided by the area of overlap between predicted bounding box and the ground-truth one. 

\subsection{Training}

The entire training is a transfer learning process with the VGG-16 feature extraction network first being pre-trained on ImageNet dataset, then MDCN is fine-tuned further on KITTI. Training was conducted on a computing cluster environment. Our models employ the training method of stochastic gradient descent (SGD)~\cite{bottou2016optimization,zhang2017bpgrad}. The momentum is set to 0.9, and the weight decay is 0.0005. The overall number of training iterations is set to 120,000. The learning rate decay policy is to maintain a constant decay factor, which multiplied the current learning rate by 0.1 at 80,000 and 100,000 iterations.

A set of default boxes do matching to the ground truth boxes, where we match each ground truth box with the best overlapped default box and those whose jaccard overlap is larger than the threshold of 0.5. MDCN imposes the set of aspect ratios for default boxes, denoted as \{1,2,3,1/2,1/3\}. We minimize the joint localization loss by Smooth L1 loss~\cite{girshick2015fast} and confidence loss by softmax loss shown in equation (6). 

\begin{equation}
\begin{aligned}
L(x,c,l,g) = \frac{1}{N}(L_{conf}+\alpha L_{loc})
\end{aligned} 
\end{equation}
where N is the number of matched default boxes and the weight term $\alpha$ is set to 1~\cite{liu2016ssd}. $L_{conf}$ and $L_{loc}$ refer to confidence loss and localization loss, respectively.

\begin{figure*}[h]
	\centering
	\includegraphics[width=0.91\linewidth]{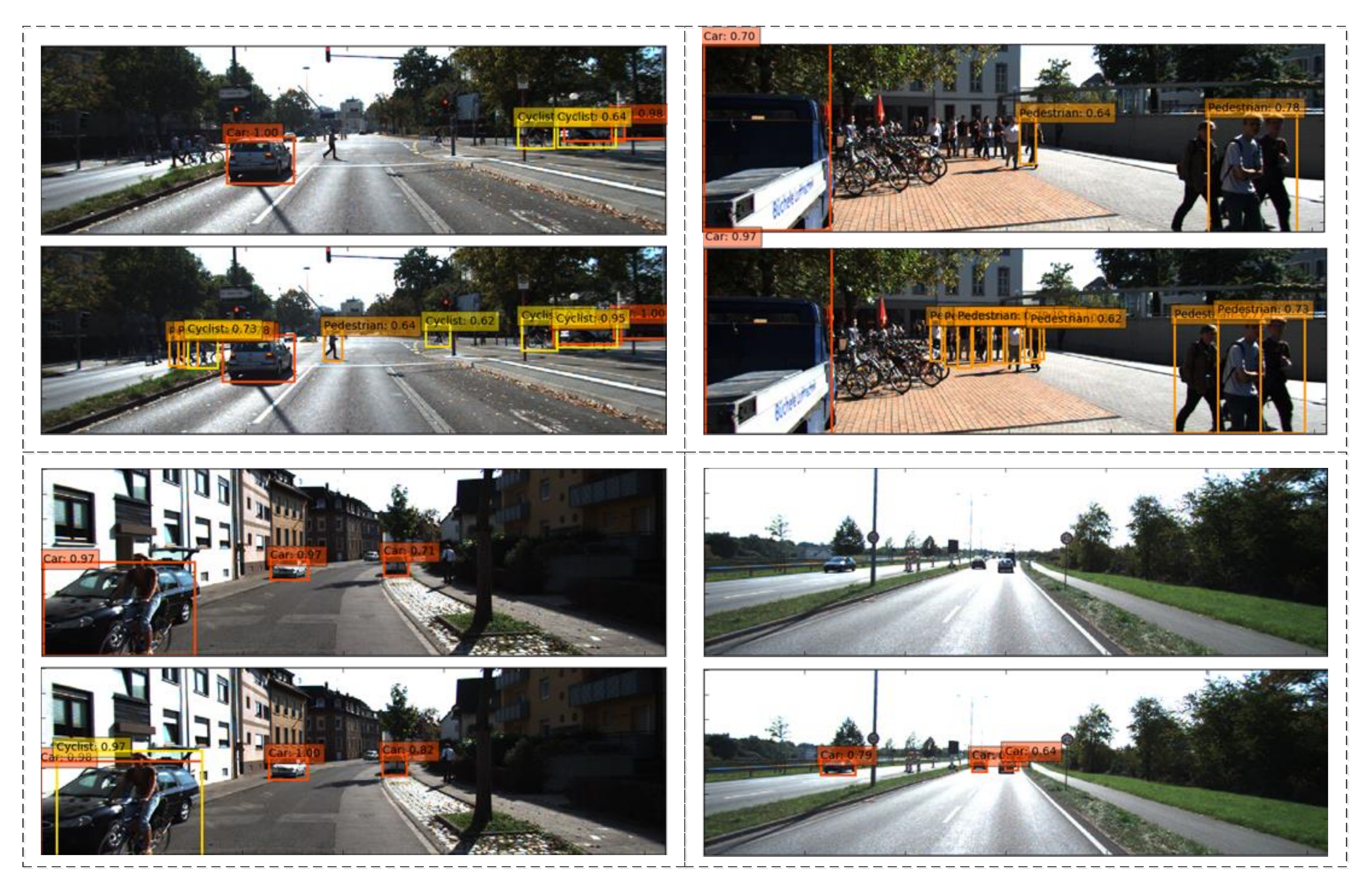}\\%
	\vspace{-3mm}
	\caption{\textbf{Detection examples of SSD and MDCN-I2.} In these four sets of images, the top one and the bottom one are from SSD and MDCN-I2, respectively.}
	\label{fig:detectionrecog2}
\end{figure*}

\subsection{Detection Accuracy}
The object detection accuracy, measured by average precision, is shown in Table \Romannum{1}. The proposed MDCN model achieved top mAP (mean average precision) on this leader board, where MDCN-I2 has 10\% higher mAP than that of WR-Inception-I2, the second leading method. MDCN takes a significant lead in all the three difficulty levels of cyclist, where MDCN-I1 even has nearly 20\% higher AP than that of the second record from WR-Inception-I2. Noticeable for car and cyclist, both MDCN models obtain better accuracies on hard level subset. Although MDCN do not perform the best in pedestrian detection, the AP of MDCN-I1 runs the first for easy objects. Since difficulty levels of objects are classified by their sizes, how much they are occluded and truncated, we can draw the conclusion MDCN performs better in the detection of small and occluded objects in complicated scenes.

Moreover, the detection accuracies of each object class under different IoU thresholds are provided in Table \Romannum{2}. MDCN obtained approximately all highest accuracies across the entire range of IoU. Especially for Car and Cyclist, MDCN-I2 gets 10\% higher AP than SSD, proving it works better when detecting hard objects and is robust in complicated scenes.    

In Fig.~\ref{fig:detectionrecog2}, the detection examples of four scenes are shown, obtained from SSD and MDCN-I2. In each group, the top image is the result of SSD, the bottom one is of MDCN-I2. MDCN-I2 obviously has a significant advantage for detecting small and occluded objects. It has stable performance for all labeled object categories. For example, in the top-left set, SSD can only detect out five objects, missing the pedestrians and cyclists around the car in the middle of the image. While for MDCN-I2, nearly all the ten objects are detected out successfully. For the top-right set, MDCN-I2 is able to detect out nearly all pedestrians even though some of them are occluded while SSD is only able to detect two pedestrians. 

\begin{table}[ht]
	\caption{Results on KITTI validation set for different IoU thresholds.} 
	\vspace{-3mm}
	\begin{center}
		\def\arraystretch{1.2}
		\scalebox{0.90}[0.90]{%
			\begin{tabular} {|c|c|c|c|c|c|c|c|c|}
				\hline
				\multirow{2}{*}{\textbf{Classes}} & \multirow{2}{*}{\textbf{Methods}} & \multicolumn{7}{c|}{\textbf{IOU}} \\ 
				\cline{3-9} 
				& & 0.5 & 0.55 & 0.6 & 0.65 & 0.7 & 0.75 & 0.8 \\
				\hline
				\multirow{3}{*}{Car} & SSD & 83.9 & 80.9 & 77.6 & 74.5 & 67.7 & 59.4 & 49.7\\ 
				\cline{2-9}
				& MDCN-I1 & 88.1 & 87.4 & 84.3 & 79.0 & \textbf{75.9} & \textbf{69.3} & 59.6\\
				\cline{2-9}
				& MDCN-I2 & \textbf{88.4} & \textbf{87.6} & \textbf{85.2} & \textbf{79.1} & \textbf{75.9} & 69.0 & \textbf{59.7}\\
				\hline
				\multirow{3}{*}{Pedestrian} & SSD & 47.3 & 41.2 & 32.7 & 27.3 & 20.8 & 15.9 & \textbf{12.4}\\
				\cline{2-9}
				& MDCN-I1 & \textbf{54.8} & \textbf{48.4} & 41.1 & 32.2 & 24.5 & \textbf{18.8} & 11.8\\
				\cline{2-9}
				& MDCN-I2 & 54.0 & 47.4 & \textbf{42.1} & \textbf{35.5} & \textbf{26.3} & 15.9 & 9.7\\
				\hline
				\multirow{3}{*}{Cyclist} & SSD & 61.5 & 52.0 & 48.7 & 41.0 & 30.2 & 21.7 & 11.0\\
				\cline{2-9}
				& MDCN-I1 & 72.8 & 62.6 & 56.9 & 51.0 & \textbf{41.0} & 28.5 & 18.1\\
				\cline{2-9}
				& MDCN-I2 & \textbf{75.0} & \textbf{68.9} & \textbf{64.3} & \textbf{52.6} & 40.1 & \textbf{28.7} & \textbf{21.8}\\
				\hline
			\end{tabular} %
		}
	\end{center}
\end{table}

\subsection{Detection Efficiency}

Due to the limitation of computation resource, we compare MDCN models with SSD, both running on GPU K40. From Table \Romannum{3}, the proposed models achieve 15-16 FPS, very close to SSD. Consider the complexity of model structures, the proposed wide-context, multi-scale inception structure does not lower the detection speed so much. This owes to the deep inception we proposed, where feature maps produced in deep layers already have much smaller sizes after down-sampling from previous layers, which dramatically decreases the computation burden for the model. Given the improvement of detection accuracy, MDCN contribute a better trade-off between accuracy and efficiency. On the other hand, MDCN-based models only surpass a little than SSD given the number of parameters. The reason lies in our proposed information square inception modules make full use of parameter-sharing and only a few layer units are introduced by multi-scale filters.

\begin{table}[ht]
	\caption{Detection efficiency of different models}
	\vspace{-3mm}
	\begin{center}
		\def\arraystretch{1.2}
		\scalebox{1.0}[1.0]{%
			\begin{tabular} {|c|c|c|c|c|c|}
				\hline
				\textbf{Model} & \textbf{Network} & \textbf{GPU} & \textbf{Resolution} & \textbf{\# of Params} & \textbf{FPS}\\
				\hline
				SSD & VGG-16 & K40 & 300$\times$300 & 2.41$\times$ 10$^{7}$ & 17.0 \\
				\hline
				MDCN-I1 & VGG-16 & K40 & 300$\times$300 & 2.54$\times$ 10$^7$ & 15.8 \\
				\hline
				MDCN-I2 & VGG-16 & K40 & 300$\times$300 & 2.55$\times$ 10$^7$ & 15.4 \\
				\hline
			\end{tabular}		
		}
	\end{center}
\end{table}

\section{Conclusion}

We have proposed the MDCN model by introducing the wide-context, multi-scale structure into a single-shot learning network. It is realized by integrating the proposed information square inception modules into the deep part of the network. The proposed framework is computational efficient with superior performance in object detection, especially for small and occluded objects. Extensive experiment on the popular KITTI dataset demonstrate the effectiveness of MDCN, which outperforms the state-of-the-art models based on single-short multi-box detector. The proposed MDCN model makes a good trade-off between efficiency and accuracy, and it is more suitable for real-world applications.  

\section*{Acknowledgment}

This work was supported in part by the Kansas NASA EPSCoR Program under Grant KNEP-PDG-10-2017-KU and the United States Department of Agriculture (USDA) under Grant USDA 2017-67007-26153. \vfill \newpage





\begin{thebibliography}{1}

\bibitem{erhan2014scalable}
D.~Erhan, C.~Szegedy, A.~Toshev, and D.~Anguelov.
\newblock Scalable object detection using deep neural networks.
\newblock In {\em CVPR}, 2014.



\bibitem{he2016deep}
K.~He, X.~Zhang, S.~Ren, and J.~Sun.
\newblock Deep residual learning for image recognition.
\newblock In {\em CVPR}, 2016.

\bibitem{lecun1998gradient}
Y.~LeCun, L.~Bottou, Y.~Bengio, and P.~Haffner.
\newblock Gradient-based learning applied to document recognition.
\newblock {\em Proceedings of the IEEE}, 86(11):1998.

\bibitem{russakovsky2015imagenet}
O.~Russakovsky, J.~Deng, H.~Su, J.~Krause, S.~Satheesh, S.~Ma, Z.~Huang,
A.~Karpathy, A.~Khosla, M.~Bernstein, et~al.
\newblock Imagenet large scale visual recognition challenge.
\newblock {\em IJCV},2015.

\bibitem{szegedy2015going}
C.~Szegedy, W.~Liu, Y.~Jia, P.~Sermanet, S.~Reed, D.~Anguelov, D.~Erhan,
V.~Vanhoucke, and A.~Rabinovich.
\newblock Going deeper with convolutions.
\newblock In {\em CVPR}, 2015.

\bibitem{zagoruyko2016wide}
S.~Zagoruyko and N.~Komodakis.
\newblock Wide residual networks.
\newblock {\em arXiv:1605.07146}, 2016.

\bibitem{huang2016densely}
G.~Huang, Z.~Liu, K.~Q. Weinberger, and L.~van~der Maaten.
\newblock Densely connected convolutional networks.
\newblock In {\em CVPR}, 2017.

\bibitem{srivastava2015training}
R.~K. Srivastava, K.~Greff, and J.~Schmidhuber.
\newblock Training very deep networks.
\newblock In {\em Advances in neural information processing systems}, 2015.

\bibitem{huang2016deep}
G.~Huang, Y.~Sun, Z.~Liu, D.~Sedra, and K.~Q. Weinberger.
\newblock Deep networks with stochastic depth.
\newblock In {\em ECCV}, 2016.

\bibitem{larsson2016fractalnet}
G.~Larsson, M.~Maire, and G.~Shakhnarovich.
\newblock Fractalnet: Ultra-deep neural networks without residuals.
\newblock {\em arXiv:1605.07648}, 2016.

\bibitem{szegedy2017inception}
C.~Szegedy, S.~Ioffe, V.~Vanhoucke, and A.~A. Alemi.
\newblock Inception-v4, inception-resnet and the impact of residual connections
on learning.
\newblock In {\em AAAI}, 2017.


\bibitem{pouyanfar2017efficient}
S.~Pouyanfar, S.-C. Chen, and M.-L. Shyu.
\newblock An efficient deep residual-inception network for multimedia
classification.
\newblock In {\em ICME}, IEEE, 2017.


\bibitem{fu2017dssd}
C.-Y. Fu, W.~Liu, A.~Ranga, A.~Tyagi, and A.~C. Berg.
\newblock Dssd: Deconvolutional single shot detector.
\newblock {\em arXiv:1701.06659}, 2017.

\bibitem{he2017mask}
K.~He, G.~Gkioxari, P.~Doll{\'a}r, and R.~Girshick.
\newblock Mask r-cnn.
\newblock In {\em ICCV}, 2017.

\bibitem{szegedy2016rethinking}
C.~Szegedy, V.~Vanhoucke, S.~Ioffe, J.~Shlens, and Z.~Wojna.
\newblock Rethinking the inception architecture for computer vision.
\newblock In {\em CVPR}, 2016.

\bibitem{geiger2012we}
A.~Geiger, P.~Lenz, and R.~Urtasun.
\newblock Are we ready for autonomous driving? the kitti vision benchmark
suite.
\newblock In {\em CVPR}, 2012.

\bibitem{ren2017accurate}
J.~Ren, X.~Chen, J.~Liu, W.~Sun, J.~Pang, Q.~Yan, Y.-W. Tai, and L.~Xu.
\newblock Accurate single stage detector using recurrent rolling convolution.
\newblock In {\em CVPR}, 2017.

\bibitem{liu2016ssd}
W.~Liu, D.~Anguelov, D.~Erhan, C.~Szegedy, S.~Reed, C.-Y. Fu, and A.~C. Berg.
\newblock Ssd: Single shot multibox detector.
\newblock In {\em ECCV}, 2016.

\bibitem{torralba2003contextual}
A.~Torralba.
\newblock Learning Depth from Single Images with Deep Neural Network Embedding Focal Length.
\newblock {\em IJCV}, 2003.


\bibitem{sui2018correlation}
Y.~Sui, G.~Wang, and L.~Z.
\newblock Correlation filter learning toward peak strength for visual tracking.
\newblock {\em IEEE transactions on cybernetics}, 2018.


\bibitem{sermanet2013pedestrian}
P.~Sermanet, K.~Kavukcuoglu, S.~Chintala, and Y.~LeCun.
\newblock Pedestrian detection with unsupervised multi-stage feature learning.
\newblock In {\em CVPR}, 2013.

\bibitem{gidaris2015object}
S.~Gidaris and N.~Komodakis.
\newblock Object detection via a multi-region and semantic segmentation-aware
cnn model.
\newblock In {\em ICCV}, 2015.

\bibitem{zagoruyko2016multipath}
S.~Zagoruyko, A.~Lerer, T.-Y. Lin, P.~O. Pinheiro, S.~Gross, S.~Chintala, and
P.~Doll{\'a}r.
\newblock A multipath network for object detection.
\newblock {\em arXiv:1604.02135}, 2016.

\bibitem{he2014spatial}
K.~He, X.~Zhang, S.~Ren, and J.~Sun.
\newblock Spatial pyramid pooling in deep convolutional networks for visual
recognition.
\newblock In {\em ECCV}, 2014.

\bibitem{iandola2016squeezenet}
F.~N. Iandola, S.~Han, M.~W. Moskewicz, K.~Ashraf, W.~J. Dally, and K.~Keutzer.
\newblock Squeezenet: Alexnet-level accuracy with 50x fewer parameters and< 0.5
mb model size.
\newblock {\em arXiv:1602.07360}, 2016.

\bibitem{jia2014caffe}
Y.~Jia, E.~Shelhamer, J.~Donahue, S.~Karayev, J.~Long, R.~Girshick,
S.~Guadarrama, and T.~Darrell.
\newblock Caffe: Convolutional architecture for fast feature embedding.
\newblock In {\em ACM}, 2014.


\bibitem{xiang2017subcategory}
Y.~Xiang, W.~Choi, Y.~Lin, and S.~Savarese.
\newblock Subcategory-aware convolutional neural networks for object proposals
and detection.
\newblock In {\em WACV}, 2017.

\bibitem{girshick2015fast}
R.~Girshick.
\newblock Fast r-cnn.
\newblock In {\em ICCV}, 2015.

\bibitem{lee2017wide}
Y.~Lee, H.~Kim, E.~Park, X.~Cui, and H.~Kim.
\newblock Wide-Residual-Inception networks for real-time object detection.
\newblock {\em IEEE Intelligent Vehicles Symposium (IV)}, 2017.

\bibitem{bottou2016optimization}
L.~Bottou, F E.~Curtis, J.~Nocedal.
\newblock Optimization methods for large-scale machine learning.
\newblock {\em arXiv:1606.04838}, 2016.

\bibitem{zhang2017bpgrad}
Z.~Zhang, Y.~Wu, and G.~Wang.
\newblock BPGrad: Towards global optimality in deep learning via branch and pruning.
\newblock {\em CVPR}, 2018.


\end{thebibliography}
%

\end{document}